\documentclass[conference]{IEEEtran}
\IEEEoverridecommandlockouts
\usepackage{cite}
\usepackage{amsmath,amssymb,amsfonts}
\usepackage{algorithmic}
\usepackage{graphicx}
\usepackage{textcomp}
\usepackage{xcolor}
\usepackage{subcaption}
\usepackage{url}
\usepackage{soul}

\def\BibTeX{{\rm B\kern-.05em{\sc i\kern-.025em b}\kern-.08em
    T\kern-.1667em\lower.7ex\hbox{E}\kern-.125emX}}

\makeatletter
\newcommand{\linebreakand}{%
  \end{@IEEEauthorhalign}
  \hfill\mbox{}\par
  \mbox{}\hfill\begin{@IEEEauthorhalign}
}
\makeatother

\begin{document}

\title{\fontsize{20pt}{20pt}\selectfont Decentralized Multi-Agent Goal Assignment for Path Planning using Large Language Models\\[0.5em]

\thanks{ This research was supported in part by Science Undergraduate Research Award (SURA) of McGill. Paper submitted at Massachusetts Institute of Technology Undergraduate Research Technology Conference (MITURTC) \\

}
}

\author{\IEEEauthorblockN{Murad Ismayilov}
\IEEEauthorblockA{\textit{Center for Intelligent Machines (CIM) Laboratory} \\
\textit{McGill University}\\
Montreal, Canada \\
Murad.Ismayilov@mail.mcgill.ca}
\and
\IEEEauthorblockN{Edwin Meriaux}
\IEEEauthorblockA{\textit{Center for Intelligent Machines (CIM) Laboratory} \\
\textit{McGill University}\\
Montreal, Canada \\
Edwin.Meriaux@mail.mcgill.ca}
\linebreakand
\IEEEauthorblockN{Shuo Wen}
\IEEEauthorblockA{\textit{Center for Intelligent Machines (CIM) Laboratory} \\
\textit{McGill University}\\
Montreal, Canada \\
Shuo.Wen@mail.mcgill.ca}
\and
\IEEEauthorblockN{Gregory Dudek}
\IEEEauthorblockA{\textit{Center for Intelligent Machines (CIM) Laboratory} \\
\textit{McGill University}\\
Montreal, Canada \\
Gregory.Dudek@mcgill.ca}
}

\maketitle

\begin{abstract}
Coordinating multiple autonomous agents in shared environments under decentralized conditions is a long-standing challenge in robotics and artificial intelligence. This work addresses the problem of decentralized goal assignment for multi-agent path planning, where agents independently generate ranked preferences over goals based on structured representations of the environment, including grid visualizations and scenario data. After this reasoning phase, agents exchange their goal rankings, and assignments are determined by a fixed, deterministic conflict-resolution rule (e.g., agent index ordering), without negotiation or iterative coordination. We systematically compare greedy heuristics, optimal assignment, and large language model (LLM)-based agents in fully observable grid-world settings. Our results show that LLM-based agents, when provided with well-designed prompts and relevant quantitative information, can achieve near-optimal makespans and consistently outperform traditional heuristics. These findings underscore the potential of language models for decentralized goal assignment in multi-agent path planning and highlight the importance of information structure in such systems.
\end{abstract}

\section{Introduction}

Coordinating the actions of multiple autonomous agents is a fundamental challenge in robotics, logistics, and artificial intelligence, with increasing relevance as multi-agent systems are deployed in real-world applications such as warehouse automation, urban delivery, and disaster response~\cite{sun2025multi, dahiya2023survey, groenewald2024multi, lundberg2009framework}. In these settings, each agent must select a unique goal or target, plan its movements, and avoid both static obstacles and conflicts with other agents operating in the same environment. Effective assignment and coordination are essential for maximizing system efficiency and avoiding deadlocks or congestion~\cite{dahiya2023survey}.

Centralized approaches to goal assignment and scheduling can, in principle, produce optimal solutions given full knowledge of the environment and control over all agents. However, such methods quickly become computationally infeasible as the number of agents grows or as the system becomes more dynamic and unpredictable~\cite{sun2025multi, sudhakara2025symmetric}.

Decentralization has emerged as a practical and scalable alternative in multi-agent systems, allowing each agent to act independently or semi-independently, often with only partial or local information about the environment and the actions of others~\cite{groenewald2024multi, sudhakara2025symmetric}. Examples include fleets of warehouse robots that must self-assign to pick locations and delivery points, or mobile robots in search-and-rescue operations, where coordination is required under uncertain and dynamic conditions~\cite{sun2025multi, dahiya2023survey}.

This work investigates the problem of decentralized goal assignment in fully observable grid-world environments. Each agent is provided with a structured representation of the world, including the grid layout, the positions of all agents, obstacles, goals, and, in some experimental conditions, explicit agent-goal distance tables. Agents independently generate a ranking of goals they prefer to pursue, based on the available information. These rankings are then collected and used to resolve assignment conflicts centrally, ensuring that each goal is assigned to exactly one agent without duplication. Notably, before assignments are finalized, agents also receive information about the provisional choices of other agents, enabling them to reason about potential conflicts and adapt their rankings accordingly.

We compare four categories of agents: a greedy baseline using nearest-goal heuristics, a centralized optimal solver and agents powered by LLMs such as GPT-4.1 and LLaVA ~\cite{liu2023visual}. LLM-based agents use structured prompts to process world information and produce their goal preferences.

By systematically evaluating these decentralized goal assignment strategies, this study aims to provide new insight into the relative strengths and limitations of algorithmic and language-based decision-making in multi-agent coordination. The findings inform the design of scalable decentralized systems, highlighting the importance of input structure, conflict resolution, and agent reasoning in collaborative environments.

\section{Background}

Classical approaches to multi-agent goal assignment have often relied on centralized planning, where a single entity with access to global information computes assignments for all agents. For example, Faigl et al.~\cite{faigl2012goal} benchmark several assignment strategies, including greedy assignment to the nearest available goal (implemented centrally to ensure unique assignment), iterative improvement, and the optimal Hungarian method, within the domain of multi-robot exploration. Their results show that methods which account for global path costs typically outperform simpler heuristics, but centralized approaches may face computational and robustness limitations in large or dynamic environments.

To address these challenges, a variety of decentralized and distributed protocols have been proposed. In decentralized settings, agents make assignment decisions using only partial knowledge or local communication, which improves robustness and scalability. Examples include auction-based algorithms, consensus-driven protocols, and hybrid metaheuristics such as the consensus-based decentralized discrete particle swarm optimization method of Tong et al.~\cite{tong2020decentralized}. These strategies have demonstrated empirical success in reducing makespan and balancing workloads among agents, especially when global knowledge is unavailable or costly to maintain.

In addition to purely algorithmic methods, recent research has explored the use of LLMs as decentralized decision-makers for multi-robot systems. For instance, Chen et al.~\cite{chen2024scalable} evaluate LLM-based planning frameworks in both centralized and decentralized multi-robot collaboration, and report that hybrid approaches leveraging real-time feedback can outperform purely centralized or decentralized strategies.

Recent advances in prompt engineering have had a notable impact on the capabilities of large language models in multi-agent coordination and planning. Prompt engineering—the process of carefully designing and refining input queries and instructions for LLMs—has been shown to significantly influence both the reliability and reasoning depth of model outputs, especially for complex decision-making and collaborative tasks. A particularly important development is chain-of-thought (CoT) prompting~\cite{wei2022chain}, where LLMs are guided to articulate intermediate reasoning steps before arriving at a final answer. CoT techniques have demonstrated substantial improvements in multi-step planning, team-based decision-making, and overall LLM performance for both individual and collaborative scenarios~\cite{chen2025unleashing}.

Empirical studies have further established that well-structured prompts—including scenario descriptions, explicit instructions, reasoning checklists, and clear conflict-resolution rules—are essential for eliciting team-level reasoning and globally efficient assignments~\cite{wei2022chain,bsharat2024principled}. These prompt design elements facilitate transparency, make agent intentions machine-interpretable, and are now considered a core component in LLM-driven multi-agent systems.

This motivates the present study, which seeks to provide a controlled evaluation of decentralized goal assignment protocols, including classical algorithms and LLM-based agents, within a unified grid-world framework.

\section{Problem Formulation}

The central problem studied in this work is decentralized goal assignment for multiple agents in a fully observable, synchronous grid-world. The objective is to assign each agent to a unique goal location so that the \textit{makespan}—the number of timesteps required for all agents to reach their assigned goals—is minimized:
\begin{equation}
\text{Makespan} = \min_{\pi} \max_{i=1, \ldots, k} C_i(\pi)
\end{equation}
where $C_i(\pi)$ is the arrival time of agent $a_i$ at its assigned goal under assignment $\pi$ and a valid set of paths.

In this problem, the algorithms only compute the assignment of agents to goals; that is, each agent is assigned to exactly one goal. Once assignments are determined, each agent takes the shortest path (by number of steps) from its initial position to its assigned goal, computed using a deterministic shortest-path algorithm such as breadth-first search (BFS)~\cite{bundy1984breadth}. No path optimization or multi-agent path planning is required beyond this assignment step.

Problems that seek to minimize makespan are instances of collaborative optimization, where agents must consider not only their own travel times but also the overall team objective. Optimal assignments often require agents to avoid purely greedy choices and instead select goals that lead to the smallest maximum arrival time across the group.

Each environment consists of:
\begin{itemize}
    \item A square $N \times N$ grid.
    \item $k$ agents, each starting from a unique cell.
    \item $k$ goals, each located in a unique cell.
    \item A set of obstacles, which are impassable cells that block agent movement.
\end{itemize}

The grid is represented as a labeled image, with obstacles shown as black squares, goals as labeled red squares (A, B, C, \ldots), agents as labeled blue circles (1, 2, 3, \ldots), and empty cells as uniquely indexed references. Borders and diagonal blockers are also depicted to aid orientation and prevent illegal moves around obstacle corners.

Agents are assumed to have full knowledge of the environment, including the grid layout, the locations of all agents and goals, and the positions of obstacles. The problem constraints are as follows:
\begin{itemize}
    \item Each goal is assigned to exactly one agent, and vice versa.
    \item Agents cannot move through obstacles, off the grid, or occupy the same cell at the same time.
    \item Agents may swap places if they simultaneously attempt to move into each other's cells.
\end{itemize}

This formulation provides a unified basis for comparing different decentralized goal assignment strategies under controlled and interpretable conditions.

\subsection{Illustrative Examples}

\begin{figure}[ht]
    \centering
    \begin{subfigure}{0.48\linewidth}
        \centering
        \includegraphics[width=1\linewidth]{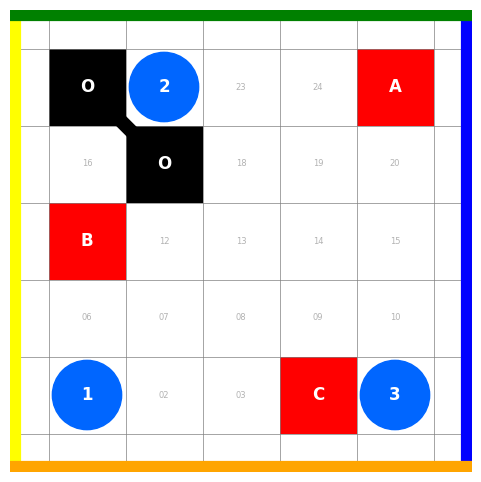}
        \caption{Initial world}
        \label{fig:config1_world}
    \end{subfigure}
    \hfill
    \begin{subfigure}{0.48\linewidth}
        \centering
        \includegraphics[width=1\linewidth]{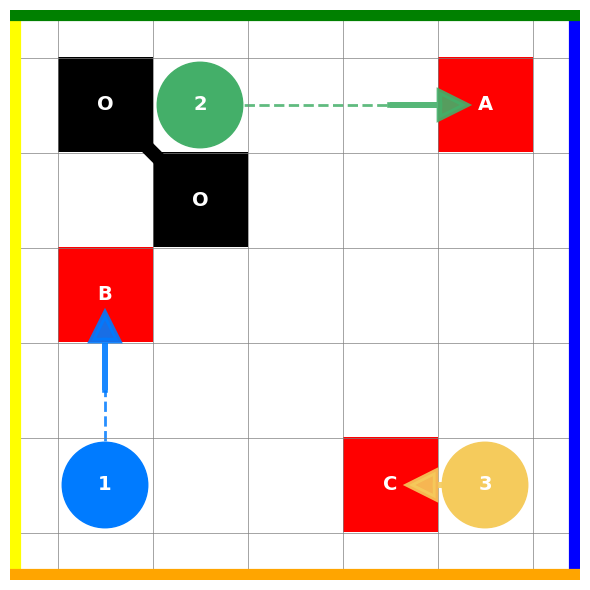}
        \caption{Optimal solution}
        \label{fig:config1_solution}
    \end{subfigure}
    \caption{Example 1 (Small World): A $5 \times 5$ grid with 3 agents, 3 goals (red color), and 2 obstacles (black color).}
    \label{fig:config1}
\end{figure}

In Example~\ref{fig:config1}, the initial world configuration (left) shows three agents and three goals in a $5 \times 5$ grid with 2 obstacles.  In this scenario, the greedy strategy (where each agent selects its nearest goal) yields the optimal assignment, as all agents can reach their respective goals along the shortest available routes without conflict. The optimal assignment, shown in the right panel, pairs Agent 1 to Goal B, Agent 2 to Goal A, and Agent 3 to Goal C. Each agent follows its shortest path to its assigned goal. The makespan in this solution—corresponding to Equation~1—is 3 steps, set by Agent 2’s route to Goal A.

\begin{figure}[ht]
    \centering
    \begin{subfigure}{0.48\linewidth}
        \centering
        \includegraphics[width=1\linewidth]{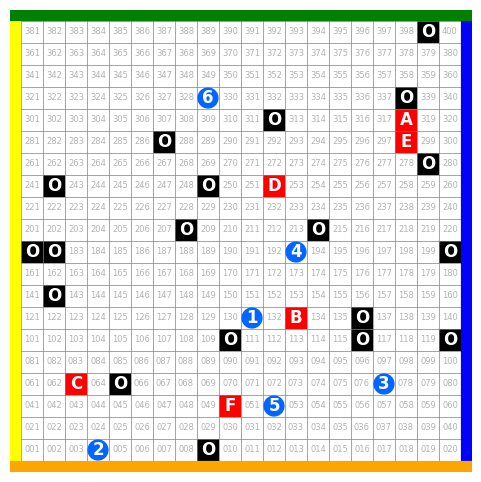}
        \caption{Initial world}
        \label{fig:config2_world}
    \end{subfigure}
    \hfill
    \begin{subfigure}{0.48\linewidth}
        \centering
        \includegraphics[width=1\linewidth]{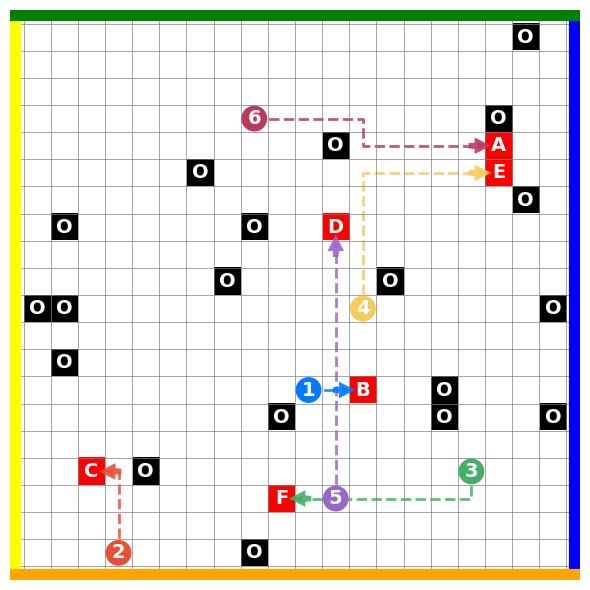}
        \caption{Optimal solution}
        \label{fig:config2_solution}
    \end{subfigure}
    \caption{Example 2 (Difficult World): A $20 \times 20$ grid with 3 agents, 3 goals, and multiple obstacles.}
    \label{fig:config2}
\end{figure}

For Example~\ref{fig:config2}, the environment contains numerous obstacles that block direct paths between agents and goals. Here, the optimal assignment may require assigning agents to non-nearest goals in order to minimize congestion and avoid bottlenecks. The increased complexity and obstacle density make it difficult to identify the optimal solution by inspection, illustrating the challenges faced by decentralized assignment strategies in large, cluttered environments.

\section{Proposed Solution}

We propose a decentralized goal assignment protocol for multi-agent systems operating in grid-based environments, leveraging large language models (LLMs) for assignment decision-making. In each scenario, agents receive a structured representation of the environment, consisting of a labeled grid image, the positions of all agents, goals, and obstacles, as well as, in some experiments, an explicit table of agent-goal distances. Based on this input, each agent independently generates a ranked list of goals according to its preferences.

Agents then simultaneously announce their ranked preferences to one another. At this point, all agents have made their decisions in a decentralized manner: no agent communicates during the decision-making process, and they only receive the goal rankings from other agents. Assignment conflicts are resolved according to a predefined agent index order. At initialization, each of the $n$ agents is assigned a unique index in the range $1$ to $n$. In the event of a conflict, the agent with the lowest index receives priority.

\subsection{Prompt Engineering}
The performance of LLM-based agents in decentralized goal assignment relies heavily on the design and structure of their prompts. In our approach, each agent receives a carefully constructed prompt that includes a labeled grid image, a scenario description, the explicit positions of all agents, goals, and obstacles, and, in some experiments, a table of agent-to-goal distances. Each prompt instructs the agent to generate a complete ranking of all goals, with explicit consideration of both team objectives and potential assignment conflicts. Guidelines for deterministic conflict resolution, such as tiebreaking by agent index, are provided to ensure agents can independently compute a consistent and feasible assignment.

A key element of our design is the use of chain-of-thought (CoT) prompting~\cite{wei2022chain}, where agents are encouraged to articulate intermediate reasoning steps before producing their final ranking. We employ explicit step-by-step checklists and structured reasoning sections in the prompt to elicit more consistent and globally informed choices from LLM agents~\cite{chen2025unleashing}.

\begin{quote}
Team-level Reasoning Checklist (excerpt):
\begin{enumerate}
    \item List every remaining goal and estimate which agent is fastest to reach each one.
    \item Draft a full assignment (agents $\rightarrow$ goals, no duplicates).
    \item Compute the assignment’s longest path length.
    \item Try at least one alternative assignment; select the one with the smallest maximum path.
    \item Try to resolve conflicts in case of ties.
\end{enumerate}
\end{quote}

The complete prompt, including the full environment specification, explicit conflict resolution logic, and formatting, is available in our project code\footnote{Full prompt and code: \url{https://github.com/MYRADhub/SURA}}.

We systematically evaluate two main prompt variants: one providing the agent-goal distance table, and one omitting this information to assess the LLM’s ability to infer costs based solely on spatial reasoning.

Additionally, our prompt protocols encode provisional assignments (“pseudo-policies”) of other agents at each step, supporting a limited form of indirect, pre-assignment communication—a design inspired by recent advances in collaborative multi-agent LLM prompting~\cite{luo2023prompt,agashe2023llmcoord}. These elements are intended to improve both single-agent accuracy and team-level coordination by making agent preferences and intentions explicit and machine-interpretable.

Our empirical results demonstrate how these prompt content and structure choices impact the effectiveness of decentralized LLM agents for multi-agent goal assignment.

\section{Methodology}

Experiments are conducted over 100 randomly generated grid-world scenarios, each consisting of a $20 \times 20$ grid with 2 to 6 agents and goals. Obstacles are placed uniformly at random, numbering between 15 and 30 per scenario, with care taken to avoid overlap with agents or goals.

To benchmark the LLM-based approach, we include several baseline strategies:
\begin{itemize}
    \item Greedy Assignment: Agents are assigned to their nearest available goal by BFS distance, with assignment order determined by agent index; assignments are final and non-negotiated.
    
    \item Random Assignment: Agents are matched to goals uniformly at random, subject to unique assignment constraints; navigation proceeds via BFS as in other methods.

\end{itemize}

In addition, for each scenario, we compute the optimal assignment via brute-force centralized search, yielding the assignment that minimizes the makespan. This provides a ground-truth lower bound for comparison, and all methods are evaluated by their absolute makespan and their performance gap relative to this optimum.

Performance is evaluated using the makespan (see Problem Formulation) and the performance gap, defined as the difference between the makespan of a given method and the optimal assignment.

\section{Results}

Table~\ref{tab:avg_makespan} reports the mean makespan achieved by each method over all scenarios. The optimal solver provides a lower bound with an average makespan of 13.93. Among LLM-based approaches, GPT-4.1 agents that re-rank goals at each step and receive explicit agent-goal distances achieve the best results, with a mean makespan of 15.12. This is closely followed by the same LLM when ranking only once at the start (15.50). When explicit distance information is removed from the prompt, GPT-4.1 performance declines to 17.67, comparable to the greedy baseline (17.93). Random assignment performs substantially worse (20.54), and the LLaVA-based LLM agent demonstrates the largest gap, with a mean makespan of 27.98.

\begin{table}[ht]
\centering
\caption{Mean makespan (timesteps until all agents reach goals) for each strategy.}
\label{tab:avg_makespan}
\begin{tabular}{lc}
\hline
\textbf{Agent Type} & \textbf{Average Makespan} \\
\hline
Optimal & 13.93 \\

\hline
LLM (rank every step, distances shown, GPT-4.1) & $\boldsymbol{15.12}$ \\
LLM (rank once, distances shown, GPT-4.1) & 15.50 \\
LLM (rank every step, no distances, GPT-4.1) & 17.67 \\
Greedy & 17.93 \\
Random Assignment & 20.54 \\
LLM (rank every step, LLaVA) & 27.98 \\
\hline
\end{tabular}
\end{table}

\begin{figure}[ht]
    \centering
    \includegraphics[width=0.9\linewidth]{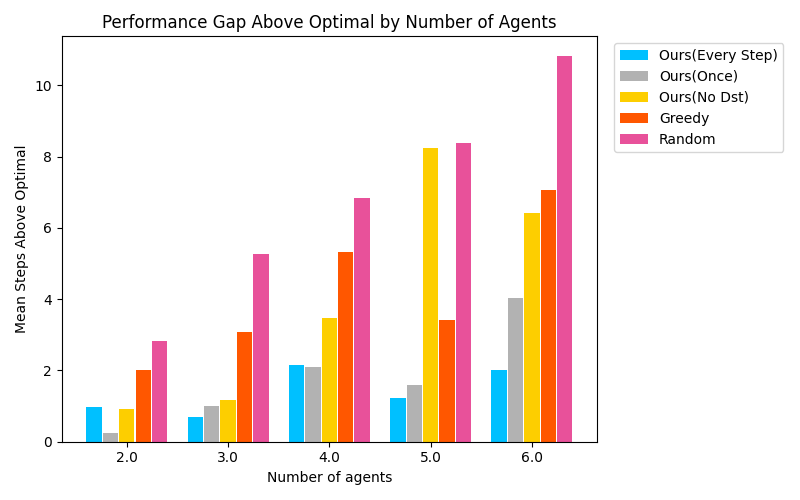}
    \caption{Performance gap (mean steps above optimal) by number of agents for each method.}
    \label{fig:gap_by_agents}
\end{figure}

To analyze how performance scales with team size, Figure~\ref{fig:gap_by_agents} plots the mean number of steps above the optimal makespan as a function of the number of agents (the lower the better). In all methods, the performance gap increases with group size, but the rate of increase varies significantly. The best LLM-based agents (GPT-4.1 with distance tables and either once or every-step ranking) remain consistently close to optimal across all scales, while LLMs deprived of distance information, greedy, and random assignment strategies show a more pronounced gap as the number of agents increases. The greedy and “no distance” LLM approaches are nearly indistinguishable for larger agent teams (except for the case with 5-agent worlds), both lagging the best LLM agents by several steps. The random assignment baseline deteriorates most rapidly as agent count grows. We can also see that ranking once strategy seems to scale worse than every-step ranking once based on the results.

These results indicate that the combination of structured world information and explicit agent-goal distances allows LLM-based agents to approximate optimal decentralized assignment even as problem complexity increases. In contrast, heuristic and unstructured methods become less effective in larger, more congested environments.

\section{Discussion}

The results of this study provide several important insights into decentralized goal assignment using LLM-based agents in grid-based environments. As expected, the optimal assignment algorithm establishes the lower bound for makespan, with all other strategies measured against this benchmark.

Among decentralized methods, LLM agents powered by GPT-4.1, given access to explicit agent-goal distance tables and updated rankings at each step, consistently achieved makespans within two steps of the optimal solution across a wide range of scenarios and team sizes. This was, on average, the best performing algorithm. It outperformed the Greedy and Random Assignment baselines, primarily because the LLM agents are able to reason about global team objectives and consider the effects of different assignments on the makespan, rather than simply selecting locally optimal or random matches. In particular, by systematically ranking assignments based on overall team cost and re-evaluating choices at each step, the LLM-based approach avoids common pitfalls of greedy assignment, such as bottlenecking a single agent or forcing suboptimal assignments due to local decisions. This solution maintained robust performance as the number of agents increased, highlighting the ability to scale to more complex coordination tasks. The single-shot ranking strategy with distance information performed only marginally worse, while depriving agents of distance tables led to a marked decline in performance—nearly matching the heuristic greedy baseline. This emphasizes that explicit, quantitative information is critical for effective decentralized assignment when using LLMs.

An analysis by group size further reinforces these findings: only the GPT-4.1 agents with distance information, along with the optimal solver, showed relatively flat scaling as agent count increased. Heuristic, random, and “no-distance” LLM strategies exhibited significantly steeper degradation, suggesting that complex environments amplify the benefits of structured input and repeated reasoning for LLMs.

In contrast, the LLaVA-based agent performed substantially worse than all other methods, with makespans often exceeding even the random assignment baseline as team size grew. This unusually poor result appears to be due in part to LLaVA's tendency to change its assignment strategy from step to step, leading to frequent shifts in goal selection and a lack of stable, globally consistent plans. On the other hand, GPT-4.1 produced more stable and coherent rankings over time, which contributed to its strong performance.

There are several limitations to the present study. All experiments were conducted in static, fully observable environments with a maximum of six agents and assumed perfect knowledge for assignment conflict resolution. Real-world scenarios may introduce dynamic obstacles, communication constraints, or partial observability that require further adaptation of these protocols. Additionally, LLM agents in this work were restricted to single-stage or per-step assignment and did not participate in navigation or ongoing negotiation.

Despite these constraints, the results demonstrate that, with well-designed prompts and access to quantitative world information, modern language models can serve as competitive decentralized agents for goal assignment.

\section{Conclusion}

This study demonstrates that large language models, when provided with structured prompts and explicit quantitative information, can serve as highly effective decentralized agents for goal assignment in multi-agent grid environments. GPT-4.1-based agents, in particular, achieved makespans close to the optimal solver without centralized planning and consistently outperformed both greedy and random assignment strategies, especially as problem complexity increased.

These results highlight the crucial role of prompt design and input structure in enhancing reasoning capabilities of LLMs for collaborative tasks. Our work provides new benchmarks for language-model-driven coordination and points to promising directions for further integrating LLM-based agents into scalable, decentralized multi-agent systems. Future work should explore larger team sizes, dynamic settings, richer agent communication protocols, and the integration of LLM reasoning into more aspects of multi-agent decision-making.

\bibliographystyle{IEEEtran}
\bibliography{references}

\begin{thebibliography}{10}
\providecommand{\url}[1]{#1}
\csname url@samestyle\endcsname
\providecommand{\newblock}{\relax}
\providecommand{\bibinfo}[2]{#2}
\providecommand{\BIBentrySTDinterwordspacing}{\spaceskip=0pt\relax}
\providecommand{\BIBentryALTinterwordstretchfactor}{4}
\providecommand{\BIBentryALTinterwordspacing}{\spaceskip=\fontdimen2\font plus
\BIBentryALTinterwordstretchfactor\fontdimen3\font minus \fontdimen4\font\relax}
\providecommand{\BIBforeignlanguage}[2]{{%
\expandafter\ifx\csname l@#1\endcsname\relax
\typeout{** WARNING: IEEEtran.bst: No hyphenation pattern has been}%
\typeout{** loaded for the language `#1'. Using the pattern for}%
\typeout{** the default language instead.}%
\else
\language=\csname l@#1\endcsname
\fi
#2}}
\providecommand{\BIBdecl}{\relax}
\BIBdecl

\bibitem{sun2025multi}
L.~Sun, Y.~Yang, Q.~Duan, Y.~Shi, C.~Lyu, Y.-C. Chang, C.-T. Lin, and Y.~Shen, ``Multi-agent coordination across diverse applications: A survey,'' \emph{arXiv preprint arXiv:2502.14743}, 2025.

\bibitem{dahiya2023survey}
A.~Dahiya, A.~M. Aroyo, K.~Dautenhahn, and S.~L. Smith, ``A survey of multi-agent human--robot interaction systems,'' \emph{Robotics and Autonomous Systems}, vol. 161, p. 104335, 2023.

\bibitem{groenewald2024multi}
C.~A. Groenewald, G.~Saha, G.~Mann, B.~Bhushan, E.~Howard, and E.~Groenewald, ``Multi-agent systems in robotics: coordination and communication using machine learning,'' \emph{Naturalista Campano}, vol.~28, pp. 882--897, 2024.

\bibitem{lundberg2009framework}
J.~Lundberg and A.~H{\aa}kansson, ``Framework for dynamic life critical situations using agents,'' in \emph{German Conference on Multiagent System Technologies}.\hskip 1em plus 0.5em minus 0.4em\relax Springer, 2009, pp. 214--219.

\bibitem{sudhakara2025symmetric}
S.~Sudhakara, ``Symmetric policy design for multi-agent dispatch coordination in supply chains,'' \emph{arXiv preprint arXiv:2504.19397}, 2025.

\bibitem{liu2023visual}
H.~Liu, C.~Li, Q.~Wu, and Y.~J. Lee, ``Visual instruction tuning,'' \emph{Advances in neural information processing systems}, vol.~36, pp. 34\,892--34\,916, 2023.

\bibitem{faigl2012goal}
J.~Faigl, M.~Kulich, and L.~Přeučil, ``Goal assignment using distance cost in multi-robot exploration,'' in \emph{2012 IEEE/RSJ International Conference on Intelligent Robots and Systems}, 2012, pp. 3741--3746.

\bibitem{tong2020decentralized}
B.~Tong, Q.~Liu, and C.~Dai, ``A decentralized hybrid method for goal assignment in multi-robot exploration,'' in \emph{2020 IEEE International Conference on Information Technology,Big Data and Artificial Intelligence (ICIBA)}, vol.~1, 2020, pp. 238--253.

\bibitem{chen2024scalable}
Y.~Chen, J.~Arkin, Y.~Zhang, N.~Roy, and C.~Fan, ``Scalable multi-robot collaboration with large language models: Centralized or decentralized systems?'' in \emph{2024 IEEE International Conference on Robotics and Automation (ICRA)}, 2024, pp. 4311--4317.

\bibitem{wei2022chain}
J.~Wei, X.~Wang, D.~Schuurmans, M.~Bosma, F.~Xia, E.~Chi, Q.~V. Le, D.~Zhou \emph{et~al.}, ``Chain-of-thought prompting elicits reasoning in large language models,'' \emph{Advances in neural information processing systems}, vol.~35, pp. 24\,824--24\,837, 2022.

\bibitem{chen2025unleashing}
B.~Chen, Z.~Zhang, N.~Langren{\'e}, and S.~Zhu, ``Unleashing the potential of prompt engineering for large language models,'' \emph{Patterns}, 2025.

\bibitem{bsharat2024principled}
S.~M. Bsharat, A.~Myrzakhan, and Z.~Shen, ``Principled instructions are all you need for questioning llama-1/2,'' GPT-3.5/4, Tech. Rep., 2024.

\bibitem{bundy1984breadth}
A.~Bundy and L.~Wallen, ``Breadth-first search,'' in \emph{Catalogue of artificial intelligence tools}.\hskip 1em plus 0.5em minus 0.4em\relax Springer, 1984, pp. 13--13.

\bibitem{luo2023prompt}
Y.~Luo, Y.~Tang, C.~Shen, Z.~Zhou, and B.~Dong, ``Prompt engineering through the lens of optimal control,'' \emph{arXiv preprint arXiv:2310.14201}, 2023.

\bibitem{agashe2023llmcoord}
S.~Agashe, Y.~Fan, A.~Reyna, and X.~E. Wang, ``Llm-coordination: evaluating and analyzing multi-agent coordination abilities in large language models,'' \emph{arXiv preprint arXiv:2310.03903}, 2023.

\end{thebibliography}

\end{document}